\documentclass[11pt, a4paper, logo, copyright, nonumbering]{map}
\usepackage{CJKutf8}
\usepackage{xargs}  
\usepackage{todonotes}
\usepackage{amsmath}
\usepackage{dsfont}

\usepackage{longtable}

\usepackage{svg}
\usepackage{fontawesome}
\usepackage{array}
\usepackage{tabularx}
\usepackage{latexsym}
\usepackage{graphicx}
\usepackage[utf8]{inputenc} 
\usepackage[utf8]{inputenc} 
\usepackage{amssymb} 
\usepackage{url}            
\usepackage{amsfonts}       
\usepackage{nicefrac}       
\usepackage{microtype}      
\usepackage{xspace}

\usepackage{listings}
\usepackage{makecell}
\usepackage{inconsolata}
\usepackage{multicol}
\usepackage{amstext}

\definecolor{darkblue}{RGB}{84, 112, 198}

\definecolor{lightblue}{rgb}{0.85, 0.95, 1.0}    
\definecolor{lightgreen}{rgb}{0.90, 1.0, 0.90}    
\definecolor{lightorange}{rgb}{1.0, 0.95, 0.85}   
\definecolor{lightpurple}{rgb}{0.95, 0.90, 1.0}   
\definecolor{lightgray}{rgb}{0.97, 0.97, 0.97}    

\usepackage{tikz}
\usetikzlibrary{calc}
\definecolor{battery-empty}{rgb}{0.9, 0.9, 0.9}
\newcommand{\difficultybar}[1]{%
  \begin{tikzpicture}[baseline, scale=0.5, every node/.style={scale=0.8}]
    \foreach \i in {1,2,3,4,5} {
      \ifnum\i>#1
        \draw[fill=battery-empty] (\i*0.5-0.5, 0) rectangle (\i*0.5, 0.25);
      \else
        \pgfmathsetmacro{\colorlevel}{80 - 12*(\i)} 
        \edef\x{\noexpand\draw[fill=blue!\colorlevel!white, opacity=0.9] (\i*0.5-0.5, 0) rectangle (\i*0.5, 0.25);}
        \x
        \draw[blue!50!black] (\i*0.5-0.5, 0) rectangle (\i*0.5, 0.25);
      \fi
    }
    \fill[battery-empty!70] (2.5, 0.08) rectangle (2.6, 0.17);
    \draw[battery-empty!70!black] (2.5, 0.08) rectangle (2.6, 0.17);
  \end{tikzpicture}%
}

\definecolor{hidden-draw}{RGB}{20,68,106}
\definecolor{hidden-pink}{RGB}{255,245,247}
\usepackage[edges]{forest}

\usepackage{natbib}
\usepackage{CJKutf8}
\usepackage{xargs}
\usepackage{todonotes}
\usepackage{multirow}
\usepackage{cleveref}
\usepackage{amsmath}
\usepackage{dsfont}
\usepackage{subcaption}
\usepackage{url}
\usepackage{svg}
\usepackage{fontawesome}
\usepackage{xcolor}
\usepackage{float}

\usepackage[utf8]{inputenc}
\usepackage{booktabs}
\usepackage{graphicx}
\usepackage{array}
\usepackage{tabularx}
\usepackage{xcolor,colortbl}  
\definecolor{boxcolor}{HTML}{d92523} 
\definecolor{bulbcolor}{HTML}{e3b87f} 
\usepackage{url}
\usepackage{ragged2e}   
\usepackage{makecell} 

\usepackage{hyperref}       
\usepackage{url}            
\usepackage{booktabs}       
\usepackage{amsfonts}       
\usepackage{nicefrac}       
\usepackage{microtype}      
\usepackage{xcolor}         
\usepackage{graphicx}
\usepackage{longtable}
\usepackage{array}
\usepackage{pbox}
\usepackage{amsmath}
\usepackage{amssymb}
\usepackage{tabularx}
\usepackage{multirow}
\usepackage{wrapfig}
\usepackage{pifont}
\usepackage{algorithm}
\usepackage{algorithmic}
\usepackage{lipsum}
\usepackage{subcaption}
\usepackage{enumitem}
\usepackage{tikz}
\usepackage{xstring}

\usepackage{color-edits}

\usepackage{threeparttable} 

\definecolor{rliableolive}{HTML}{BBCC33}
\definecolor{rliableblue}{HTML}{77AADD}
\definecolor{rliablered}{HTML}{f63c44}

\usepackage[most,skins,theorems]{tcolorbox}
\tcbuselibrary{skins,breakable,fitting,hooks}  
\tcbset{
  aibox/.style={
    width=\linewidth,
    top=7pt,
    bottom=2pt,
    colback=rliablered!18!white,
    colframe=black,
    colbacktitle=black,
    enhanced,
    center,
    attach boxed title to top left={yshift=-0.1in,xshift=0.15in},
    boxed title style={boxrule=0pt,colframe=white,},
  }
}
\newtcolorbox{AIbox}[2][]{aibox,title=#2,#1}

\addauthor[Zhejian]{zzj}{blue}

\hypersetup{
    colorlinks=true,            
    linkcolor=blue,             
    filecolor=magenta,          
    urlcolor=cyan,              
    citecolor=purple,             
    pdftitle={Overleaf Example},
    pdfpagemode=FullScreen,
}

\renewcommand{\today}{}
\newcommandx{\info}[2][1=]{\todo[linecolor=red,backgroundcolor=red!25,bordercolor=red,#1]{#2}}

\title{\vspace{-0.2in}
\centering \fontsize{15pt}{16pt}\selectfont Scaling Laws for Code: Every Programming Language Matters
\vspace{-0.2in}
}

\author{
\centering
\textbf{Jian Yang}\textsuperscript{1} 
\textbf{Shawn Guo}\textsuperscript{2} 
\textbf{Lin Jing}\textsuperscript{2} 
\textbf{Wei Zhang}\textsuperscript{1}
\textbf{Aishan Liu}\textsuperscript{1}  
\textbf{Chuan Hao}\textsuperscript{2}  
\\
\textbf{Zhoujun Li}\textsuperscript{1}
Wayne Xin Zhao\textsuperscript{3}
Xianglong Liu\textsuperscript{1$\dagger$} 
Weifeng Lv\textsuperscript{1$\dagger$} 
Bryan Dai\textsuperscript{2} \\
\textsuperscript{1}Beihang University
\textsuperscript{2}Ubiquant \\
\textsuperscript{3}Gaoling School of Artificial Intelligence, Renmin University of China\\
\textsuperscript{$\dagger$}Corresponding Authors. Email: \texttt{\{xlliu, lwf\}@buaa.edu.cn}
\vspace{-5pt}
}

\begin{abstract}

\vspace{-0.2in}

Code large language models (Code LLMs) are powerful but costly to train, with scaling laws predicting performance from model size, data, and compute.
However, different programming languages (PLs) have varying impacts during pre-training that significantly affect base model performance, leading to inaccurate performance prediction. 
Besides, existing works focus on language-agnostic settings, neglecting the inherently multilingual nature of modern software development.
Therefore, it is first necessary to investigate the scaling laws of different PLs, and then consider their mutual influences to arrive at the final multilingual scaling law.
In this paper, we present the first systematic exploration of scaling laws for multilingual code pre-training, conducting over 1000+ experiments (Equivalent to 336,000+ H800 hours) across multiple PLs, model sizes (0.2B to 14B parameters), and dataset sizes (1T tokens). We establish comprehensive scaling laws for code LLMs across multiple PLs, revealing that interpreted languages (e.g., Python) benefit more from increased model size and data than compiled languages (e.g., Rust). The study demonstrates that multilingual pre-training provides synergistic benefits, particularly between syntactically similar PLs. Further, the pre-training strategy of the parallel pairing (concatenating code snippets with their translations) significantly enhances cross-lingual abilities with favorable scaling properties. Finally, a proportion-dependent multilingual scaling law is proposed to optimally allocate training tokens by prioritizing high-utility PLs (e.g., Python), balancing high-synergy pairs (e.g., JavaScript-TypeScript), and reducing allocation to fast-saturating languages (Rust), achieving superior average performance across all PLs compared to uniform distribution under the same compute budget.

\end{abstract}

\begin{document}

\maketitle

\let\oldthefootnote\thefootnote


\begin{figure*}[h!]
    \centering
    \includegraphics[width=1.0\textwidth]{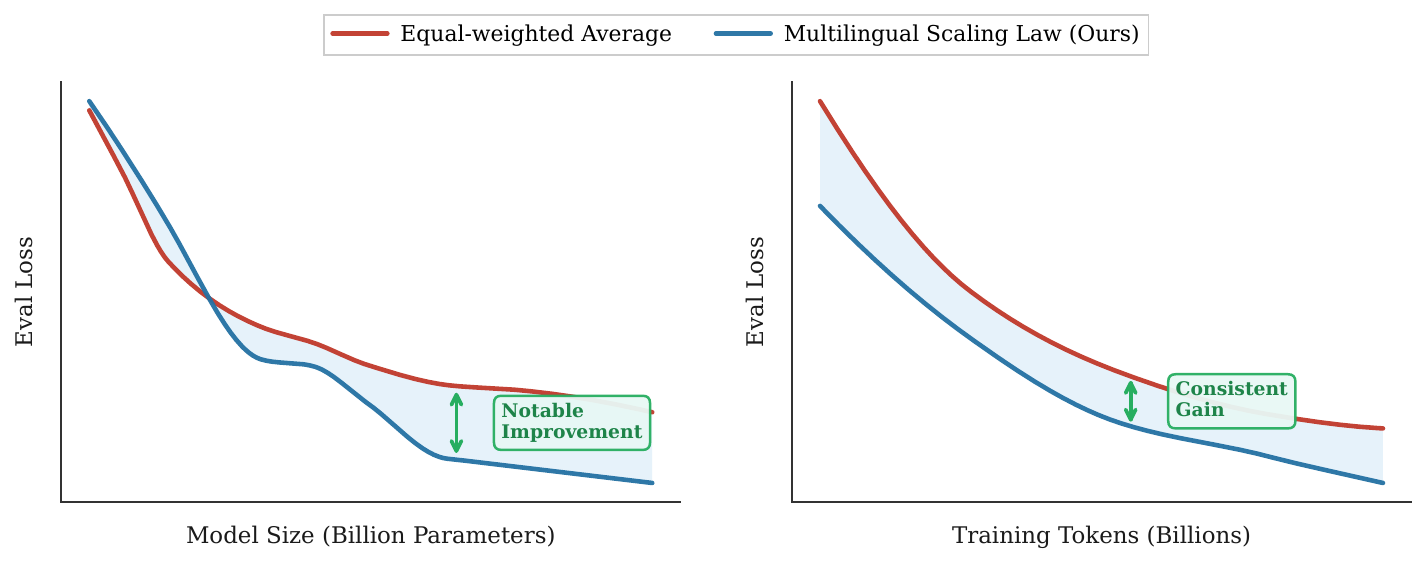}
    \caption{ Evaluation loss comparison showing that the proposed multilingual scaling law achieves consistently lower loss than the baseline across model sizes and token budgets.}
    \label{fig:intro}
\end{figure*}

\section{Introduction}
Code large language models (code LLMs) have achieved excellent coding performance in multiple programming languages (PLs), guided by the scaling law in general domains~\cite{codex,qwen25coder,deepseek_coder,guiagents_foundation_models,deepseekv1,gpt4o}. Code LLMs significantly enhance developer productivity~\cite{cursor2025features}, but training top-tier LLMs consumes enormous computing resources and costs~\cite{scaling_laws_lm,chinchilla,gpt3,deepseekv1}, making it prohibitively expensive to conduct ablation experiments on data composition (e.g., proportions of different PLs) or pre-training strategies at scale. This cost barrier limits our ability to systematically understand the key factors driving code LLM performance and hinders the development of more efficient training methodologies.

Scaling laws in general domains present a principled approach to characterizing how model performance (validation loss) depends on essential variables, such as model size, dataset size, and compute budget~\cite{scaling_laws_lm,chinchilla}. The recent work~\cite{code_scaling_law} has extended scaling laws from natural language to code, revealing that code pre-training is significantly more data-hungry compared to standard pre-training, which requires larger data-to-parameter ratios to achieve optimal performance. Yet modern code pre-training software development is inherently multilingual, with developers routinely working across Python, Java, JavaScript, C++, and numerous other PLs for downstream applications and tasks (e.g., OSAgents~\cite{osagents} and TerminalBench~\cite{terminalbench}). This reality raises critical unanswered questions: \textit{(1) What is the scaling law for multilingual code pre-training and multilingual cross-lingual capabilities? (2) What is the optimal strategy for allocating training resources across different PLs?}

Understanding multilingual scaling dynamics is crucial for several reasons. First, real-world code repositories exhibit substantial linguistic diversity, with languages varying in syntactic structure, semantic properties, and data availability. Second, multilingual pretraining may enable cross-lingual transfer, where knowledge from one language improves performance on others. Third, programming languages universally share core semantic concepts for manipulating data and controlling logic, even though their surface-level syntax and expressive paradigms can differ dramatically. Despite the practical importance of these questions, existing scaling law research focuses exclusively on monolingual or language-agnostic settings, leaving the multilingual dimension unexplored.

To address this gap, we conduct the first systematic exploration of scaling laws for multilingual code pretraining. Our study comprises over 1000+ experiments spanning multiple PLs, model sizes (0.2B to 14B parameters), and dataset sizes (1T tokens), where all base models are pre-trained from scratch. In Figure~\ref{fig:intro}, our multilingual scaling law consistently outperforms the equal-weighted baseline across different model scales and training regimes. We investigate four fundamental aspects: (1) the trade-off between language diversity and language depth, (2) cross-lingual transfer effects in multilingual pre-training, (3) scaling dynamics for code translation tasks, and (4) optimal language proportion allocation strategies. Our findings reveal insights about the interplay between linguistic diversity and model performance, providing actionable guidance for training multilingual code LLMs.

Our key contributions include:
\begin{itemize}
    \item We establish language-specific scaling laws for each PLs independently, revealing that interpreted PLs (e.g., Python) show larger scaling exponents (benefiting more from increased model size and data) compared to compiled PLs (e.g., Rust). The irreducible loss metric establishes a complexity ordering (C\# $<$ Java $\approx$ Rust $<$ Go $<$ TypeScript $<$ JavaScript $<$ Python), where PLs with strict syntax are more learnable and predictable than dynamically-typed PLs.
    \item We study synergy gain matrix of different PLs, showing that PLs with similar syntax or structure  (e.g., Java-C\#) can bring positive transfer compared to single-PL pre-training. Most PLs can benefit from multilingual pre-training, indicating that optimal PL mixing strategies in pre-training must be tailored to the characteristics of each PL.
    \item This research investigates how data organization strategies during pre-training affect cross-lingual abilities of code LLMs. The experiments discover that parallel pairing, which concatenates code snippets with their translations to provide explicit document-level alignment, significantly outperforms baseline on multilingual code translation and generation. The parallel pairing strategy exhibits favorable scaling laws, with a high scaling exponent, enabling efficient utilization of model capacity to learn cross-lingual alignments of different translation directions.
    \item We propose a proportion-dependent multilingual scaling law that incorporates language-specific parameters $(\alpha_N, \alpha_D, L_{\infty})$ and the parameter of cross-lingual transfer effects $\tau_{ij}$. Under optimal allocation, the model allocates more tokens to high-utility languages (e.g., Python) and balanced amounts to high-synergy pairs (e.g., JavaScript-TypeScript) while reducing tokens for fast-saturating languages (e.g., Rust). Evaluation of the multilingual code generation demonstrates that optimized allocation achieves higher average performance across all PLs without significant degradation in any single language, validating that strategic reallocation based on scaling laws and language synergies outperforms uniform distribution under identical compute budgets.
\end{itemize}

\section{Scaling Laws for Code Pre-training}
\subsection{ChinChilla Scaling Law for Code}
Scaling laws provide a theoretical framework for understanding how model performance evolves with computational resources. For LLM, the relationship between validation loss $\mathcal{L}$ and key factors (model parameters $N$, training tokens $D$, and compute budget $C$) can be characterized by power-law formulations~\cite{chinchilla} as below:
\begin{MiddleEquation}
\begin{equation}
\mathcal{L}(N, D) = \left(\frac{N_c}{N}\right)^{\alpha_N} + \left(\frac{D_c}{D}\right)^{\alpha_D} + \mathcal{L}_{\infty}
\end{equation}
\end{MiddleEquation}where $\mathcal{L}$ is the cross-entropy loss, $N$ is the number of model parameters, $D$ is the dataset size (number of tokens), and $C$ is the compute budget (FLOPs). $N_c$ and $D_c$ are scaling constants. $\alpha_{N}$ and $\alpha_{D}$ are power law exponents, and $L_{\infty}$ is the irreducible loss (the irreducible loss represents the inherent error that no model can eliminate, regardless of its size or training quality.
). 

\begin{figure*}[t!]
    \centering
    \includegraphics[width=1.0\textwidth]{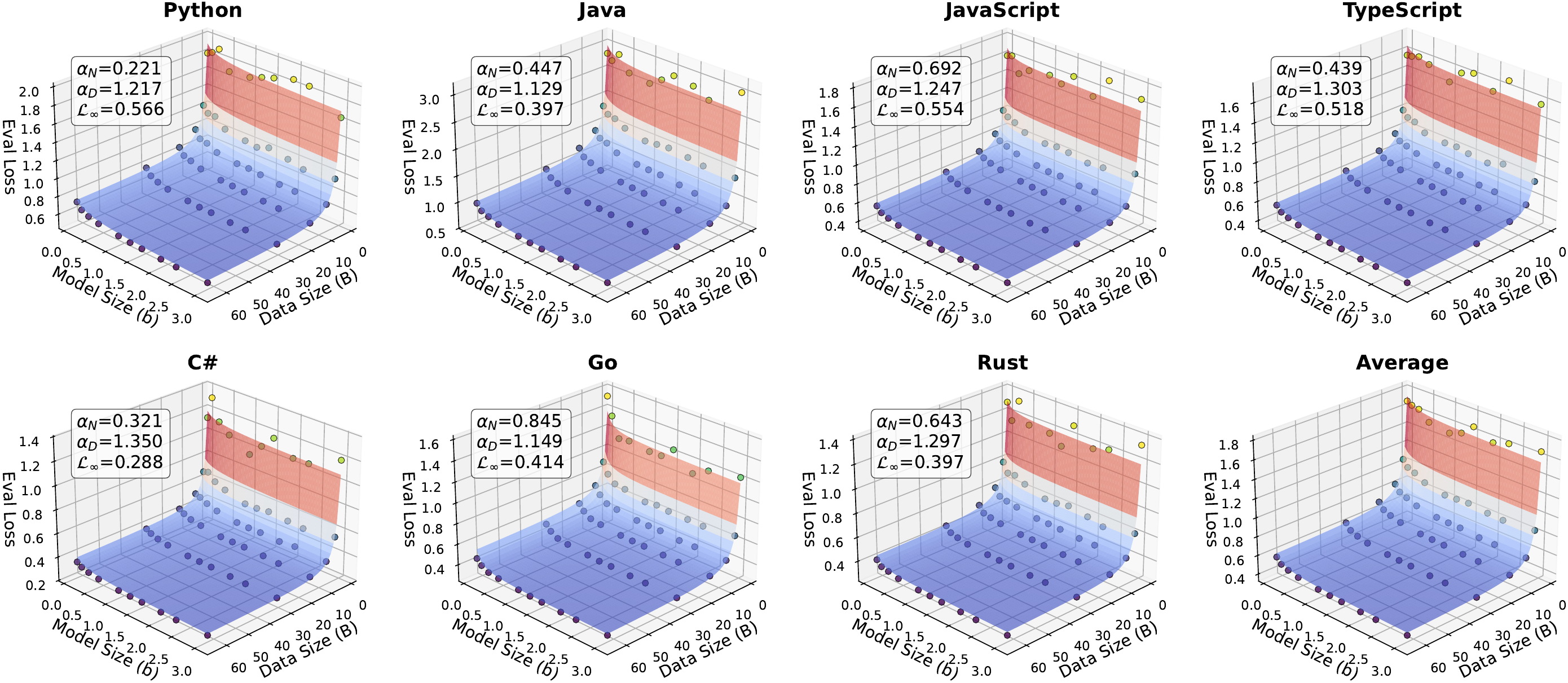}
    \vspace{-5mm}
    \caption{Scaling Laws for each PL independently. It shows a clear ordering of intrinsic predictability across PLs: C\# $<$ Java $\approx$ Rust $<$ Go $<$ TypeScript $<$ JavaScript $<$ Python.}
    \label{fig:scaling_laws_for_each_pl}
\end{figure*}
\subsection{Motivation and Formulation}
Our goal is to address three fundamental questions about multilingual code pre-training through a systematic empirical investigation:

\textbf{(1) Language-Specific Scaling Dynamics:} How does each PL scale with model size and data? We aim to investigate whether different languages exhibit distinct scaling exponents and irreducible losses, thereby challenging the assumption that all programming languages can be treated uniformly in pre-training.

\textbf{(2) Cross-Lingual Synergy Effects:} What are the synergistic or antagonistic effects when mixing different PLs during pre-training? We investigate whether bilingual training can outperform monolingual baselines under fixed compute budgets and identify which language combinations yield positive transfer versus harmful interference.

\textbf{(3) Compositional Cross-Lingual Transfer:} Can LLMs trained on supervised language pairs generalize to unseen language pairs? We examine whether different data organization strategies (from random shuffling to supervised pair training) enable zero-shot transfer between languages without direct parallel data.

To address these questions, we conduct three complementary experimental studies spanning over 1000+ training runs across diverse model scales (0.2B to 14B parameters) and data volumes (up to 1T tokens). Our approach isolates specific variables while controlling for confounding factors, enabling the derivation of actionable scaling laws for practical multilingual code pre-training.

\section{Language-specific Scaling Law}
\begin{AIbox}{\textsc{Takeaways}: Language-specifc Scaling Law}
\begin{itemize}[itemsep=2pt]
\setlength{\leftskip}{-20pt}
    \item For code across all PLs, scaling up data size yields greater performance gains than scaling up model size.
    \item Different PLs exhibit distinct convergence rates during training.
    \item PLs vary significantly in their inherent training difficulty.
\end{itemize} 
\end{AIbox}

\subsection{Experimental Setting}
We first try to establish baseline scaling laws for each programming language in isolation, investigating whether different programming languages exhibit distinct scaling exponents ($\alpha_N$, $\alpha_D$). and how the irreducible loss $L_{\infty}$ varies across languages and what this reveals about intrinsic language complexity. We select $7$ PLs that span diverse paradigms and application domains, including Python, Java, JavaScript, TypeScript, C\#, Go, and Rust. For each PL, we train LLMs with different model parameters (different model sizes of 10 settings: 0.1B, 0.2B, 0.4B, 0.6B, 1.1B, 1.3B, 1.6B, 2.0B, 2.4B, 3.1B) and token budgets (budget training tokens of 6 settings: 2B, 4B, 8B, 16B, 32B, and 64B tokens), yielding $10$ (\text{model sizes}) $\times$ $6$ (\text{token budgets}) $\times$ $7$ (\text{PLs}) = $420$ experiments in total. These experiments provide dense coverage of the $(N, D)$ space, enabling robust estimation of Chinchilla scaling law formulations for each PL.

To ensure a fair comparison across languages, all LLMs 
share the same model architecture, similar to LLaMA-2, including SwiGLU activations, rotary position embeddings (RoPE), multi-head attention (MHA), and RMSNorm . We collect a high-quality training corpus spanning multiple languages, where Python and other PLs are parallel corpus.



\subsection{Language-Specific Scaling Laws}
To establish the scaling laws for each PL, we train 420 LLMs across 7 PLs with systematic variations in model size ($N$) and training tokens ($D$). For each language, we fit both the Chinchilla-style power law.

\paragraph{Scaling Exponents and Optimal Allocation}
\autoref{fig:scaling_laws_for_each_pl} summarizes the fitted scaling parameters for each PL. The results reveal significant heterogeneity in scaling behaviors across PLs, challenging the assumption that all programming languages can be treated uniformly. The results show a clear pattern that interpreted languages exhibit larger scaling exponents than compiled languages. Python, as a dynamically-typed interpreted PL, obtains the highest $\alpha_N$ and $\alpha_D$ values, indicating that it benefits more aggressively from increases in both model parameters and training data. In contrast, Rust, as a statically-typed compiled PL with strict memory safety guarantees, shows notably smaller exponents. The explicit type annotations and rigid syntactic structure of Rust make it more learnable for LLMs with fewer parameters and less training data compared to dynamically-typed languages. The optimal $N$:$D$ ratios also vary substantially across languages, reflecting different compute-data trade-offs. PLs with higher $\alpha_D$ relative to $\alpha_N$ (e.g., Python) favor larger training datasets for a given parameter budget, while PLs with lower $\alpha_D$ (e.g., Rust) can achieve comparable performance with relatively fewer tokens but may benefit from increased model capacity.


\paragraph{Irreducible Loss and Language Complexity}
The irreducible loss $\mathcal{L}_{\infty}$ serves as a fundamental measure of language complexity, where the lower bound on achievable perplexity given infinite compute. Our results show a clear ordering: C\# $<$ Java $\approx$ Rust $<$ Go $<$ TypeScript $<$ JavaScript $<$ Python, providing insights into the intrinsic predictability of different programming languages. C\# achieves the lowest $\mathcal{L}_{\infty}$ due to its strict type system, consistent naming conventions, and standardized ecosystem. Java and Rust both enforce strong syntactic and semantic constraints that limit expression diversity. The minimalist design of Go3 yields moderate predictability, while TypeScript retains partial unpredictability from JavaScript through optional typing. The high $\mathcal{L}_{\infty}$ of JavaScript arises from its dynamic typing, flexible paradigms, and lack of uniform standards. Python exhibits the highest $\mathcal{L}_{\infty}$, reflecting its dynamic and expressive nature, diverse idioms, and variability in coding style.

\section{Language Mixture for Data Scarcity}
\begin{AIbox}{\textsc{Takeaways}: Language-specifc Scaling Law}
\begin{itemize}[itemsep=2pt]
\setlength{\leftskip}{-20pt}
    \item For code data across all programming languages, scaling up data size yields greater performance gains than scaling up model size.
    \item Different PLs exhibit distinct convergence rates during training.
    \item PLs vary significantly in their inherent training difficulty.
\end{itemize} 
\end{AIbox}
\subsection{Experimental Setting}
The second experiment investigates the effects of mixing two PLs during pre-training, examining whether mixing with a different PL improves performance compared to pre-training on a single PL. We use the total training budget of 128B tokens and compare different data compositions for each PL. For a given target PL $L_{i}$ (e.g., Python), we construct a baseline training setting by repeating $D_{L_{i}}$ ($|D_{L_{i}}|= 64$ B) twice and obtain a total of 128B tokens. Then, we mix the PL $L_{i}$ ($|D_{L_{i}}| = 64$B) and PL $L_{j}$ ($|D_{L_{j}}|=64$B) to study the language interference between $L_{i}$ and $L_{j}$. In summary, we compare pre-training setting (1): $D_{L_{i}} + D_{L_{i}}$ and setting (2): $D_{L_{i}} + D_{L_{j}}$. After pre-training, LLM is only evaluated on downstream tasks in the target language $L_{i}$ to study the interference of $L_{j}$ to $L_{i}$. For example, an LLM trained on ``Python + Java'' is evaluated on Python validation loss. 
This setup allows us to quantify the benefit (or harm) of including auxiliary language data. We define the synergy gain as:
\begin{MiddleEquation}
\begin{equation}
\Delta(L_{i}, L_{j}) = \mathcal{L}(L_{i} + L_{j}) - \mathcal{L}(L_{i} +  L_{i})
\end{equation}
\end{MiddleEquation}where $\mathcal{L}(L_{i} + L_{j})$ denotes the validation loss trained on $D_{L_{i}} + D_{L_{j}}$ while $\mathcal{L}(L_{i} + L_{i})$ denotes the validation loss trained on $D_{L_{i}} + D_{L_{i}}$. A positive $\Delta$ indicates that mixing with PL $L_{j}$ improves performance on $L_{i}$ compared to the self-repetition baseline. For most low-resource PLs, providing auxiliary PLs can significantly enhance low-resource language performance.

\subsection{Bilingual Mixture Effects}
We train 28 LLMs with different mixing PLs ($\frac{7 \times 6}{2} + 7 = 28$) with a fixed architecture ranging from 0.1B to 3.1B parameters and 128B total tokens.

\paragraph{Synergy Gain Matrix}
\autoref{tab:synergy_matrix_reordered} presents the synergy gain $\Delta(\mathcal{L}_{i}, \mathcal{L}_{j})$ for all language pairs. A positive value indicates that PL $L_{i}$ mixing with the auxiliary PL $L_{j}$ improves performance on the target PL $L_{i}$ compared to the self-repetition baseline ($D_{L_{i}} \times 2$). The diagonal entries are zero by definition (self-repetition baseline). Our results reveal two key findings: (1) The multilingual pre-training provides substantial benefits for most PLs, with 6/7 languages showing consistent positive synergy across all auxiliary language combinations. (2) PLs that share similar syntax, semantics, or programming paradigms benefit significantly from joint training, as the LLM can leverage shared statistical patterns and transfer learned representations across languages. Java exhibits exceptional synergy gains with all auxiliary PLs. Mixing Java with C\# yields the largest improvement ($\Delta = 0.186$), followed by JavaScript ($\Delta = 0.114$), TypeScript ($\Delta = 0.109$), and Rust ($\Delta = 0.112$). This suggests that Java, despite its mature ecosystem, benefits enormously from exposure to diverse programming paradigms during pre-training. The Java-C\# pair likely benefits from their shared object-oriented paradigm, similar standard libraries, and common design patterns.

We also observe that certain language pairs outperform self-repetition baselines. The Java-C\# combination achieves validation loss of 0.718 compared to 0.903 for Java self-repetition---a remarkable 20.5\% improvement. Similarly, JavaScript-TypeScript mixing (0.517 vs 0.542) and TypeScript-JavaScript mixing (0.517 vs 0.535) both outperform their respective self-repetition baselines. This suggests that exposure to a related but distinct language introduces beneficial regularization or diversifies the learned representations in a way that substantially improves generalization.

However, when Python is the target PL, mixing with most auxiliary PL produces small negative effects: JavaScript ($\Delta = -0.009$), TypeScript ($\Delta = -0.007$), C\# ($\Delta = -0.013$), Go ($\Delta = -0.016$), and Rust ($\Delta = -0.021$). Only Java provides a modest positive synergy ($\Delta = 0.010$). The cross-lingual transfer is asymmetric: mixing Python as an auxiliary language benefits other target languages (e.g., $\Delta = 0.054$ for Java, $\Delta = 0.030$ for JavaScript), but Python itself suffers when mixed with them. Overall, negative interference is limited and language-specific rather than a general phenomenon. The results indicate that multilingual pre-training benefits most programming languages, though Python may require tailored mixture strategies.

\begin{table*}[t]
\centering
\definecolor{improvement}{HTML}{D7191C} 
\definecolor{decline}{HTML}{008888}    

\resizebox{1.0\textwidth}{!}{%
\begin{tabular}{c|ccccccc}
\toprule
\textbf{Language} & \textbf{Python} & \textbf{Java} & \textbf{JavaScript} & \textbf{TypeScript} & \textbf{C\#} & \textbf{Go} & \textbf{Rust} \\
\midrule

Python 
    & 0.7528 
    & \cellcolor{improvement!15} 0.7600 (\textcolor{improvement}{$\uparrow$}\textbf{1.36\%}) 
    & \cellcolor{decline!10} 0.7733 (\textcolor{decline}{$\downarrow$}\textbf{1.12\%})     
    & \cellcolor{decline!10} 0.7426 (\textcolor{decline}{$\downarrow$}\textbf{0.95\%})     
    & \cellcolor{decline!15} 0.7688 (\textcolor{decline}{$\downarrow$}\textbf{1.69\%})     
    & \cellcolor{decline!20} 0.7613 (\textcolor{decline}{$\downarrow$}\textbf{2.13\%})     
    & \cellcolor{decline!25} 0.7656 (\textcolor{decline}{$\downarrow$}\textbf{2.72\%})     
    \\
\addlinespace[3pt]

Java 
    & \cellcolor{improvement!45} 0.8490 (\textcolor{improvement}{$\uparrow$}\textbf{6.02\%}) 
    & 0.7942 
    & \cellcolor{improvement!70} 0.7913 (\textcolor{improvement}{$\uparrow$}\textbf{12.62\%}) 
    & \cellcolor{improvement!70} 0.9034 (\textcolor{improvement}{$\uparrow$}\textbf{12.08\%}) 
    & \cellcolor{improvement!85} 0.8069 (\textcolor{improvement}{$\uparrow$}\textbf{20.58\%}) 
    & \cellcolor{improvement!65} 0.7894 (\textcolor{improvement}{$\uparrow$}\textbf{10.68\%}) 
    & \cellcolor{improvement!70} 0.7175 (\textcolor{improvement}{$\uparrow$}\textbf{12.41\%}) 
    \\
\addlinespace[3pt]

JavaScript 
    & \cellcolor{improvement!45} 0.5126 (\textcolor{improvement}{$\uparrow$}\textbf{5.49\%}) 
    & \cellcolor{improvement!25} 0.5170 (\textcolor{improvement}{$\uparrow$}\textbf{2.98\%}) 
    & 0.5285 
    & \cellcolor{improvement!40} 0.5262 (\textcolor{improvement}{$\uparrow$}\textbf{4.69\%}) 
    & \cellcolor{improvement!20} 0.5352 (\textcolor{improvement}{$\uparrow$}\textbf{2.44\%}) 
    & \cellcolor{improvement!12} 0.5424 (\textcolor{improvement}{$\uparrow$}\textbf{1.34\%}) 
    & \cellcolor{improvement!22} 0.5292 (\textcolor{improvement}{$\uparrow$}\textbf{2.56\%}) 
    \\
\addlinespace[3pt]

TypeScript 
    & \cellcolor{improvement!35} 0.5124 (\textcolor{improvement}{$\uparrow$}\textbf{4.17\%}) 
    & \cellcolor{improvement!25} 0.5347 (\textcolor{improvement}{$\uparrow$}\textbf{2.29\%}) 
    & \cellcolor{improvement!30} 0.5284 (\textcolor{improvement}{$\uparrow$}\textbf{3.34\%}) 
    & 0.5225 
    & \cellcolor{improvement!15} 0.5273 (\textcolor{improvement}{$\uparrow$}\textbf{1.68\%}) 
    & \cellcolor{improvement!15} 0.5169 (\textcolor{improvement}{$\uparrow$}\textbf{1.39\%}) 
    & \cellcolor{improvement!10} 0.5257 (\textcolor{improvement}{$\uparrow$}\textbf{1.18\%}) 
    \\
\addlinespace[3pt]

C\# 
    & \cellcolor{improvement!30} 0.3327 (\textcolor{improvement}{$\uparrow$}\textbf{3.84\%}) 
    & \cellcolor{improvement!15} 0.3331 (\textcolor{improvement}{$\uparrow$}\textbf{1.93\%}) 
    & \cellcolor{improvement!25} 0.3391 (\textcolor{improvement}{$\uparrow$}\textbf{3.10\%}) 
    & \cellcolor{improvement!30} 0.3393 (\textcolor{improvement}{$\uparrow$}\textbf{3.71\%}) 
    & 0.3395 
    & \cellcolor{improvement!15} 0.3352 (\textcolor{improvement}{$\uparrow$}\textbf{1.87\%}) 
    & \cellcolor{improvement!15} 0.3459 (\textcolor{improvement}{$\uparrow$}\textbf{1.98\%}) 
    \\
\addlinespace[3pt]

Go 
    & \cellcolor{improvement!40} 0.4121 (\textcolor{improvement}{$\uparrow$}\textbf{4.77\%}) 
    & \cellcolor{improvement!25} 0.4137 (\textcolor{improvement}{$\uparrow$}\textbf{2.95\%}) 
    & \cellcolor{improvement!22} 0.4204 (\textcolor{improvement}{$\uparrow$}\textbf{2.70\%}) 
    & \cellcolor{improvement!35} 0.4200 (\textcolor{improvement}{$\uparrow$}\textbf{4.41\%}) 
    & \cellcolor{improvement!25} 0.4328 (\textcolor{improvement}{$\uparrow$}\textbf{2.95\%}) 
    & 0.4211 
    & \cellcolor{improvement!25} 0.4200 (\textcolor{improvement}{$\uparrow$}\textbf{2.86\%}) 
    \\
\addlinespace[3pt]

Rust 
    & \cellcolor{improvement!30} 0.3801 (\textcolor{improvement}{$\uparrow$}\textbf{3.87\%}) 
    & \cellcolor{improvement!25} 0.3794 (\textcolor{improvement}{$\uparrow$}\textbf{2.89\%}) 
    & \cellcolor{improvement!35} 0.3954 (\textcolor{improvement}{$\uparrow$}\textbf{4.20\%}) 
    & \cellcolor{improvement!35} 0.3840 (\textcolor{improvement}{$\uparrow$}\textbf{4.05\%}) 
    & \cellcolor{improvement!25} 0.3844 (\textcolor{improvement}{$\uparrow$}\textbf{2.81\%}) 
    & \cellcolor{improvement!25} 0.3788 (\textcolor{improvement}{$\uparrow$}\textbf{2.80\%}) 
    & 0.3843 
    \\
\bottomrule
\end{tabular}%
}
\vspace{-2mm}
\caption{Synergy gain matrix (Reordered). Values indicate absolute performance, while parentheses show relative improvement vs baseline. \textbf{Bold numbers} indicate the percentage change. Background intensity indicates magnitude (Darker Red = Higher Gain).}
\label{tab:synergy_matrix_reordered}
\end{table*}

\paragraph{Suggestions for Data Curation}
These findings have direct implications for constructing multilingual code corpora. When training tokens are limited, prioritize mixing syntactically-related PLs can further bring more improvement compared to naively upsampling a single PL. The positive synergy effects suggest that linguistic diversity, particularly when it spans across the code domain, acts as a form of data augmentation that improves model robustness. For realistic multilingual pre-training, a mixed-language training regime is superior to language-specific fine-tuning. However, the optimal mixing ratio remains an open question. While the experiments of this section use a 50:50 ratio for language mixing, \autoref{guideline_for_multilingual_code_pretraining} will investigate whether asymmetric mixtures (e.g., 75:25) yield further improvements.

\section{Cross-Lingual Scaling Laws}
\subsection{Experimental Setting}
\begin{AIbox}{\textsc{Takeaways}: Cross-Lingual Pre-training Strategies}
\textbf{Data:} Parallel data for \texttt{Python}$\leftrightarrow$\texttt{Others} translation pairs, plus monolingual data for all 7 PLs.
\begin{enumerate}[label=\textbf{(\arabic*)}, itemsep=3pt, leftmargin=10pt]
    \item \textbf{Implicit:} Train on shuffled monolingual data; \, Test on all directions
    \item \textbf{Supervised:} Train on \texttt{Python}$\leftrightarrow$\texttt{Others}; \, Test on \texttt{Python}$\leftrightarrow$\texttt{Others}
    \item \textbf{Zero-Shot:} Train on \texttt{Python}$\leftrightarrow$\texttt{Others}; \, Test on \texttt{Non-Python} pairs (e.g., \texttt{Java}$\to$\texttt{C\#})
\end{enumerate}
\end{AIbox}
\paragraph{Pre-training Strategies} This section investigates how different pre-training strategies affect cross-lingual transfer capabilities. 
Given $K$ PLs, there are $n = K \times (K-1)$ translation directions in total. We only have Python-centric training parallel data ($t=2K$ translation directions), while the other $m = n - t$ directions have no data.
(1) The LLM is pre-trained on $\{D_{{1}},\dots,D_{{K}}\}$ ($D_{k}$ denotes the dataset of PL $L_{k}$), and then we evaluate the average translation loss of different PLs on \{$L_{i_{1}\to j_{1}},\dots,L_{i_{t}\to j_{t}}$\} ($L_{i_1\to j_1}$ denotes the translation direction from PL $L_{i_1}$ to $L_{j_1}$)
(2) The LLM is pre-trained on $t$ translation direction $\{D_{i_{1}\to j_{1}},\dots,D_{i_{t}\to j_{t}}\}$, and then we evaluate the average translation loss on $t$ translation direction $\{L_{i_{1}\to j_{1}},\dots,L_{i_{t}\to j_{t}}\}$.
(3) The LLM is pre-trained on Python-centric parallel multilingual data $\{D_{L_{i_{1}\to j_{1}}},\dots,D_{i_{t}\to j_{t}}\}$, and then we evaluate the average translation loss on $m$ unseen translation directions \{$L_{i_{t+1}\to j_{t+1}},\dots,L_{i_{n}\to j_{n}}$\}.
\paragraph{Training Corpus} We construct a comprehensive multilingual corpus of 900B tokens containing algorithmically equivalent implementations across seven programming languages, where Python serves as a pivot language with parallel implementations to six target languages (Java, JavaScript, TypeScript, C\#, Go, and Rust). Notably, direct pairs between non-Python languages are absent from the training data. We augment this with 100B tokens from FineWeb-Edu for natural language understanding, yielding 1T total tokens. We train LLMs with different model parameters (different model sizes of 10 settings: 0.1B, 0.2B, 0.4B, 0.6B, 1.1B, 1.3B, 1.6B, 2.0B, 2.4B,
3.1B) and token budgets (budget training tokens of 6 set-
tings: 2B, 4B, 8B, 16B, 32B, and 64B tokens)
We systematically evaluate two data organization paradigms across five model scales (0.2B, 0.5B, 1.5B, 3B, and 7B parameters), training each configuration for a full epoch over the 1T token corpus, comprised of 900B code tokens and 100B natural language tokens for natural language understanding.
\paragraph{Cross-lingual Evaluation}
To assess cross-lingual capabilities, we construct a held-out evaluation set for PL translation task through a careful curation process. Three software engineers select 50 Python files from GitHub, ensuring that each code snippet is functionally translatable to all target PLs and the samples span diverse algorithmic tasks to avoid evaluation bias toward specific programming paradigms. Human annotators then manually produce equivalent implementations for $6$ target PLs (Java, JavaScript, TypeScript, C\#, Go, and Rust), following strict guidelines to preserve semantic equivalence while adhering to language-specific idioms. The resulting evaluation set comprises $50 \times A_7^2 = 2{,}100$ translation instances covering all $42$ translation directions, with an average sequence length of $464$ tokens. This comprehensive coverage enables systematic evaluation of both seen translation directions (Python $\leftrightarrow$ Others) and unseen zero-shot directions (Non-Python $\leftrightarrow$ Non-Python). Given the source code $x$ of PL $L_{i}$, we calculate the loss $-\mathbb{E}[\log{P(y|x)]}$ of the target code $y$ of PL $L_{j}$.

\subsection{Pre-training on Unsupervised Multilingual Data}
\begin{table*}[t]
\centering
\resizebox{\textwidth}{!}{
\begin{tabular}{@{}lccccccc@{}}
\toprule
 & Python & Java & Go & C\# & Javascript & Typescript & Rust \\
\midrule
Python & -- & 0.71, 0.87, 0.11, 0.25 & 0.56, 1.45, 0.20, 0.28 & 0.62, 1.36, 0.19, 0.29 & 0.68, 1.45, 0.14, 0.38 & 0.61, 1.70, 0.19, 0.36 & 0.38, 2.40, 0.36, 0.46 \\
Java & 0.80, 0.89, 0.07, 0.36 & -- & 0.69, 1.22, 0.15, 0.30 & 0.53, 2.92, 0.22, 0.54 & 0.37, 1.35, 0.28, 0.46 & 0.59, 1.11, 0.16, 0.35 & 0.43, 1.45, 0.28, 0.21 \\
Go & 0.18, 0.72, 0.38, 0.31 & 0.61, 0.80, 0.08, 0.17 & -- & 0.53, 0.89, 0.15, 0.16 & 0.70, 0.81, 0.11, 0.24 & 0.10, 8.31, 0.72, 1.05 & 0.04, 0.65, 1.24, 0.39 \\
C\# & 0.19, 0.90, 0.38, 0.09 & 0.51, 1.85, 0.15, 0.46 & 0.81, 1.73, 0.12, 0.52 & -- & 0.14, 5.73, 0.68, 0.89 & 0.16, 5.15, 0.74, 0.85 & 0.04, 4.70, 1.32, 0.87 \\
Javascript & 0.47, 0.77, 0.17, 0.11 & 0.63, 0.74, 0.10, 0.19 & 0.61, 1.35, 0.17, 0.28 & 0.55, 1.06, 0.19, 0.23 & -- & 0.33, 11.88, 0.47, 0.98 & 0.30, 1.04, 0.31, 0.10 \\
Typescript & 0.58, 0.82, 0.12, 0.36 & 0.81, 1.33, 0.08, 0.51 & 0.63, 1.01, 0.14, 0.24 & 0.23, 0.95, 0.49, 0.10 & 0.56, 40.06, 0.21, 1.20 & -- & 0.11, 1.20, 0.79, 0.08 \\
Rust & 0.56, 0.81, 0.13, 0.14 & 0.56, 1.07, 0.14, 0.20 & 0.68, 1.07, 0.14, 0.27 & 0.43, 1.02, 0.17, 0.14 & 0.57, 0.86, 0.16, 0.17 & 0.27, 2.40, 0.31, 0.66 & -- \\
\bottomrule
\end{tabular}}
\caption{Chinchilla scaling law parameters $(A, B,\alpha_N, \alpha_D)$ for baseline model across translation directions. Formula: $\mathcal{L}(N,D) =  A/N^{\alpha_N} + B/D^{\alpha_D} + L_{\infty}$.}
\label{tab:chinchilla_scalinglaw_matrix}
\end{table*}
\paragraph{Zero-shot Scaling Law}
Since the standard pre-training provides no direct alignment between language pairs, we evaluate LLM on code translation loss that require implicit cross-lingual capabilities. The zero-shot scaling follows:
\begin{MiddleEquation}
\begin{equation}
\mathcal{L}_{z}(N) = A_{z} \cdot N^{-\alpha_{z}} + B_{z} \cdot D^{-\beta{z}} +\mathcal{L}_{\infty,z}
\end{equation}
\end{MiddleEquation}where $A_{z} = 0.1574$,$B_{z} = 9.553$, $\alpha_{z} = 0.3470$,$\beta_{z} = 0.8829$, and $L_{\infty,z} = 0.1236$. Even without explicit supervision, larger LLMs develop emergent cross-lingual capability, suggesting that pre-training on unsupervised multilingual code data can gets the basic cross-lingual capability.

\subsection{Pre-training on Supervised Multilingual Data}
Unlike pre-training without the explicit cross-lingual alignment, we concatenate 
the code snippet $x$ of PL $L_{i}$ and the corresponding translation $y$ of PL $L_{j}$ as $(x, y)$, which provides an explicit document-level alignment signal. 

\paragraph{Translation Scaling Law}
For language pairs explicitly aligned during training (Python $\leftrightarrow$ \{Java, JavaScript, TypeScript, C\#, Go, Rust\}), we observe enhanced cross-lingual performance that scales as:
\begin{MiddleEquation}
\begin{equation}
\mathcal{L}_{a}(N) = A_{a} \cdot N^{-\alpha_{a}} +  B_{a} \cdot D^{-\beta_{a}}+\mathcal{L}_{\infty,a}
\label{eq:supervised_code_translation}
\end{equation}
\end{MiddleEquation}where $A_{a} =0.0508$,$B_{a} = 0.793$, $\alpha_{a} = 6.404$,$\beta_{a} = 0.8829$, and $L_{\infty,a} = 0.1006$. where the high scaling exponent $\alpha_z$ indicates that parallel pairing enables efficient exploitation of model capacity for learning cross-lingual alignment between seen language pairs.

\paragraph{Zero-shot Translation Scaling Law}
Critically, document-level pairing also improves zero-shot performance on unseen language pairs (e.g., Java $\leftrightarrow$ Go, Rust $\leftrightarrow$ JavaScript). Despite never observing direct alignments between non-Python languages, models exhibit compositional generalization:
\begin{MiddleEquation}
\begin{equation}
\mathcal{L}_{zt}(N) = A_{zt} \cdot N^{-\alpha_{zt}} +  B_{zt} \cdot D^{-\beta_{zt}}+\mathcal{L}_{\infty,zt}
\label{eq:zero_shot_supervised_code_translation}
\end{equation}
\end{MiddleEquation}where $A_{zt} = 0.0350$, $B_{zt} = 4.518$, $\alpha_{zt} = 0.781$, $\beta_{zt} = 0.869$, and $L_{\infty,zt} = 0.0524$. The equation reveal that zero-shot performance under document-level pairing substantially exceeds that of the shuffled baseline. This suggests that the model leverages Python as an implicit bridge, composing translations through learned bidirectional mappings (e.g., Java $\rightarrow$ Python $\rightarrow$ Go). \autoref{tab:chinchilla_scalinglaw_matrix} summarizes the fitted scaling parameters across strategies and evaluation settings, demonstrating that parallel pairing yields superior scaling performance for both seen and unseen translation diretions.

\subsection{Cross-Lingual Translation Strategies}
This subsection examines how different data organization strategies during pre-training affect the ability of LLM to perform cross-lingual code translation. We compare two strategies, including (1) random shuffling and (2) parallel pairing, across five LLM sizes (0.2B to 7B parameters) on 1T tokens.

\begin{figure}[htbp]
    \centering
    \includegraphics[width=0.95\textwidth]{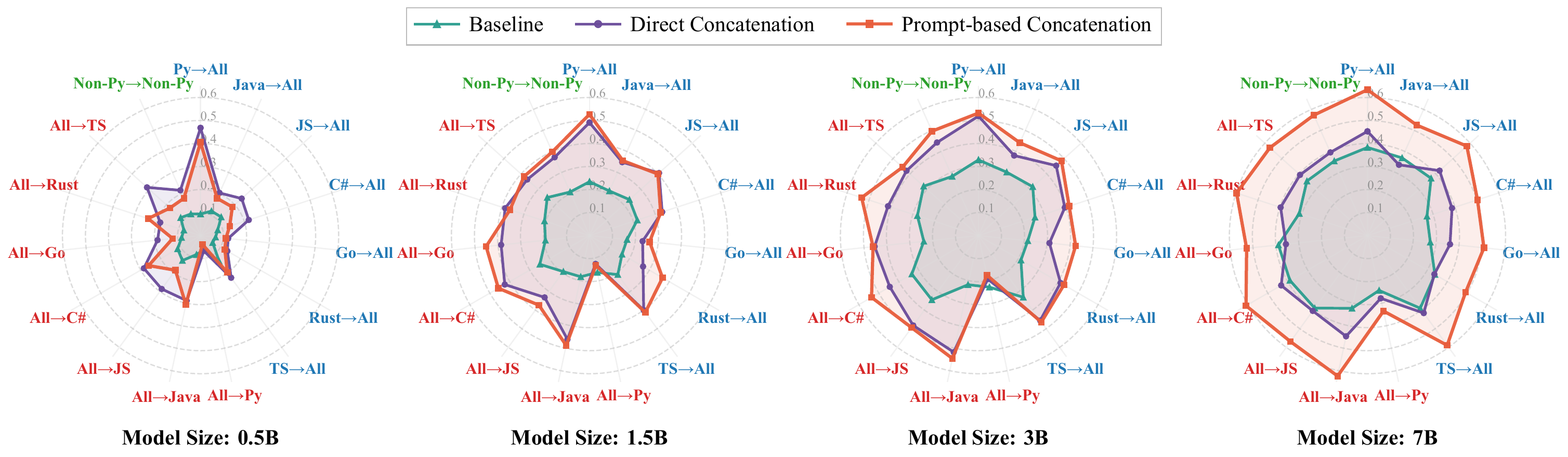}
    \caption{Translation scores across 3 strategies, 7 programming languages (PLs), and 42 directions. We aggregated the results by averaging based on language and direction into three categories: from each language to others, from others to a specific language, and between other languages, excluding Python. Across different model sizes, both \textbf{prompt-based concatenation} and \textbf{direct concatenation} significantly outperform the Baseline. Furthermore, we observe that scores for translations from other languages to Python are significantly lower than for other directions.}
    \label{tab:translation_all}
\end{figure}

\paragraph{Performance on Code Translation}
\autoref{tab:translation_all} reports the average validation loss on the 12 seen translation directions (Python $\leftrightarrow$ \{Java, JavaScript, TypeScript, C\#, Go, Rust\}). As expected, the parallel pairing strategy achieves better performance compared to the random shuffling strategy. The parallel pairing acts as a soft alignment signal to enforce the LLM to learn the cross-lingual alignment. To further evaluate the cross-lingual capability, we evaluate all LLMs on multilingual code generation. 
\autoref{tab:translation_all} explores two strategies of code concatenation, including direct concatenation ($x + y$) and prompt-based concatenation ($x + y$).

\paragraph{Performance on Code Generation} \autoref{tab:general_benchmarks} presents the evaluation results on the multilingual code generation benchmark MultiPL-E.
The LLM trained with parallel pairing also gets the better multilingual code generation performance 
This suggests that document-level pairing is the optimal data organization strategy for multilingual code pre-training, balancing translation competence with general-purpose code understanding.

\begin{table}[t]
\centering
\resizebox{0.65\columnwidth}{!}{
\begin{tabular}{lccccc}
\toprule
\textbf{PL} & \textbf{0.5B} & \textbf{1.5B} & \textbf{3B} & \textbf{7B}  \\
\midrule
Python     & 14.02 / 12.20 & 19.51 / 21.34  & 21.95 / 25.61 & 34.15 / 26.22  \\
Java       & 1.90 / 8.23   & 5.70 / 15.84  & 6.96 / 22.78  & 14.56 / 32.28  \\
JavaScript & 7.45 / 7.45   & 21.74 / 19.89 & 24.22 / 24.84 & 36.02 / 37.27  \\
TypeScript & 15.72 / 13.21 & 22.64 / 21.93 & 29.56 / 28.30 & 40.25 / 40.25  \\
C\#        & 10.13 / 9.49  & 15.19 / 13.35 & 25.32 / 24.05 & 37.34 / 32.91  \\
\midrule
Average    & 9.84 / 10.12  & 16.96 / 18.47 & 21.60 / 25.12 & 32.46 / 33.79  \\
\bottomrule
\end{tabular}}
\caption{Evaluation results on multilingual code generation benchmark MultiPl-E, Baseline VS Doc Level}
\label{tab:general_benchmarks}
\end{table}

\paragraph{Zero-Shot Translation on Unseen Directions}
A key question is whether LLMs can generalize to translation directions not seen during pre-training, particularly between non-Python language pairs (e.g., Java $\to$ Go, Rust $\to$ JavaScript). \autoref{tab:translation_all} reports validation loss on the 30 unseen translation directions. Both strategies demonstrate zero-shot translation capability on unseen directions. For the random shuffling strategy, it performs poorly on seen directions, but does not completely fail on unseen pairs, which can generate syntactically plausible translations, albeit with higher error rates. This suggests that the model learns some general notion of algorithmic equivalence across languages, even without explicit alignment during training.
LLMs trained on parallel PLs achieve better performance on unseen directions compared to the baseline.
\begin{figure}[htbp]
    \centering
    \includegraphics[width=0.65\columnwidth]{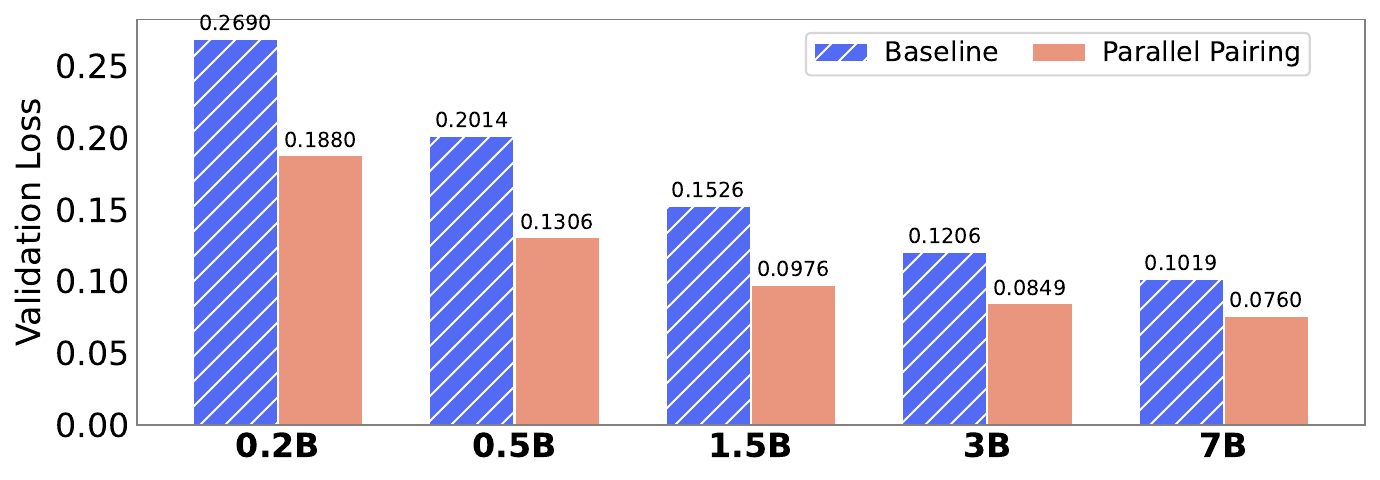}
    \vspace{-2mm}
    \caption{Validation loss on unseen translation directions (non-Python language pairs). Each entry is the average loss across 30 translation pairs not seen during pre-training.}
    \label{tab:translation_unseen}
\end{figure}

\paragraph{Scaling Laws for Code Translation}
To quantify how translation performance scales with model size and data, we fit power-law curves to the validation loss on both seen and unseen translation directions. In \autoref{eq:supervised_code_translation} and \autoref{eq:zero_shot_supervised_code_translation}, the fitted exponents $\alpha$ reveal how efficiently each strategy leverages additional model capacity. Strategies with higher $\alpha$ benefit more from scaling, while lower $L_{\infty}$ indicates better asymptotic performance. These scaling laws enable practitioners to predict translation quality at scales beyond what was experimentally evaluated, informing decisions about model size and training compute allocation.

\section{Guideline for Code Pre-training}
\label{guideline_for_multilingual_code_pretraining}
\begin{AIbox}{\textsc{Takeaways}: Cross-Lingual Pre-training Strategies}
\begin{enumerate}[label=\textbf{(\arabic*)}, itemsep=3pt, leftmargin=10pt]
    \item Uniform allocation is suboptimal (even modest adjustments yield measurable gains).
    \item Language synergy effects are substantial and should guide corpus design.
    \item The methodology can generalize beyond seven codes to any multilingual code pre-training setting.
\end{enumerate}
\end{AIbox}
\subsection{Experimental Setup}
\begin{figure}[htbp]
    \centering
    \includegraphics[width=0.75\columnwidth]{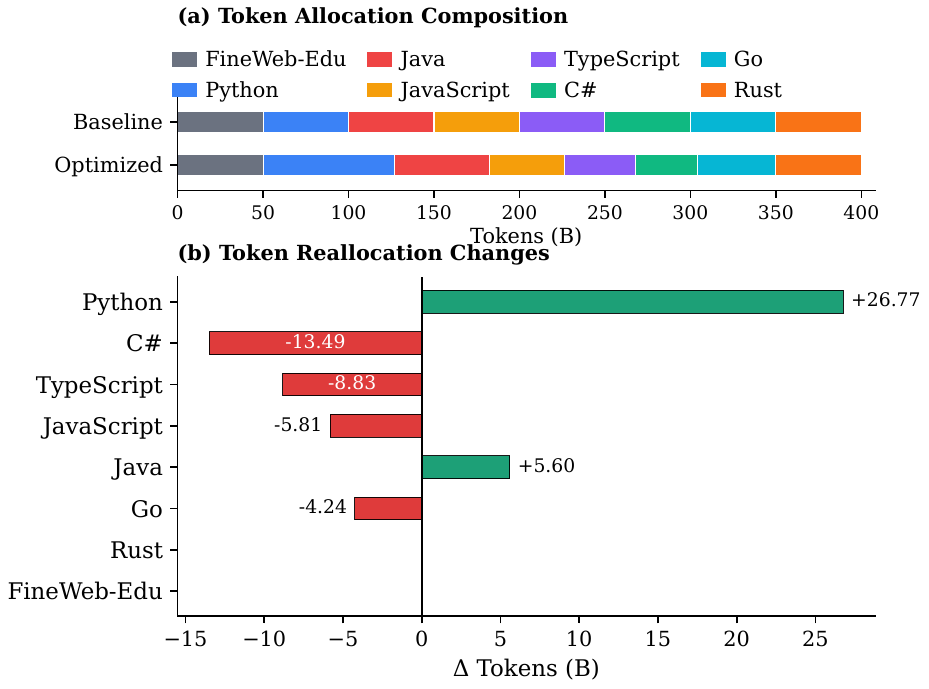}
    \vspace{-2mm}
\caption{Token allocation comparison between baseline and optimized strategies. Both use 350B total code tokens but with different distributions. The optimized allocation is derived from fitted scaling laws ($\alpha_N$, $\alpha_D$, $L_{\infty}$), optimal $N$:$D$ ratios, and synergy gain analysis.}
\label{tab:token_allocation}
\end{figure}

We train two 1.5B parameter models with different training data distributions (400B tokens: 350B code + 50B FineWeb-Edu): (1) \textbf{Baseline (Uniform Allocation):} Equal allocation of 50B tokens to each PL (350B code + 50B FineWeb-Edu = 400B total), representing standard multilingual pre-training practice. (2) \textbf{Optimized (Guided Allocation):} Strategic allocation of the same 350B code tokens based on fitted scaling laws, synergy matrix, and language complexity analysis (350B code + 50B FineWeb-Edu = 400B total). \autoref{tab:token_allocation} presents the detailed token distribution for both strategies. The optimized allocation redistributes tokens based on marginal utility: more for high-$\alpha_D$ languages (Python), balanced allocation for high-synergy pairs (Java-C\#, JavaScript-TypeScript), and reduced tokens for fast-saturating languages (e.g, Go).

\subsection{Proportion-dependent Multilingual Scaling Law}

Traditional scaling laws treat multilingual code as homogeneous, but PLs contribute differently to performance. We extend this by incorporating language proportions $p
= (p_1, \ldots, p_K)$ explicitly:
\begin{equation}
\mathcal{L}(N, D; p) = A \cdot N^{-\alpha_N(p)} + B \cdot D_{x}^{-\alpha_D(p)} + L_\infty(p)
\end{equation}
where $\alpha_N(p) = \sum_{k} p_{k} \alpha_N^k$, $\alpha_D(p) = \sum_{k} p_{k} \alpha_D^{k}$, and $L_\infty(p) = \sum_{p} p L_\infty^k$ are proportion-weighted averages of language-specific parameters from \autoref{fig:scaling_laws_for_each_pl}. The effective data term captures the effects of the cross-lingual transfer:

\begin{MiddleEquation}
\begin{equation}
D_{x} = D_{all} \left(1 + \gamma \sum_{L_{i} \neq L_{j}} p_{L_{i}} p_{L_{j}} \tau_{ij} \right)
\end{equation}
\end{MiddleEquation}where $\tau_{ij}$ is the transfer coefficient derived from \autoref{tab:synergy_matrix_reordered}. 

\subsection{Scaling Law under Optimal Allocation}

Substituting the optimal proportions $p*$ from \autoref{tab:token_allocation}:
\begin{MiddleEquation}
\begin{equation}
\mathcal{L}^*(N, D) = A^* \cdot N^{-\alpha^*_N} + B^* \cdot D^{-\alpha_D^*} + L_{\infty}^*
\end{equation}
\end{MiddleEquation}where $\alpha_D^* = 0.6859$, $\alpha_N^* = 0.2186$, $L_{\infty}^* = 0.2025$ are the fitted parameters under the optimal multilingual allocation for the multilingual code generation and translation at the same time. These coefficients are obtained through weighted fitting of the multilingual code generation and the translation loss.

\subsection{Evaluation on Multilingual Code Generation and Translation}

\begin{figure}[htbp]
    \centering
    \begin{subfigure}[b]{0.49\textwidth}
        \centering
        \includegraphics[width=\linewidth]{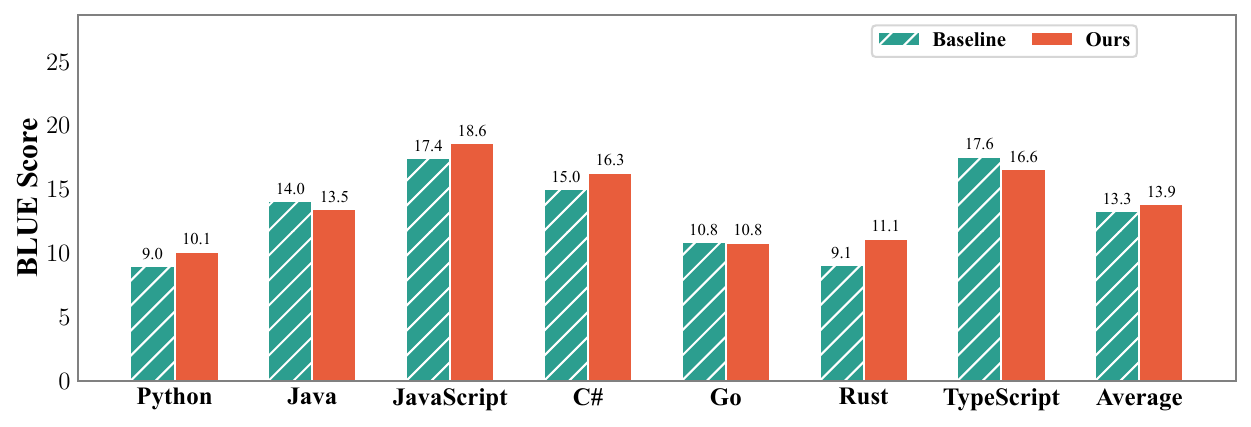}
        \caption{Multilingual translation performance}
        \label{fig:left_image}
    \end{subfigure}
    \hfill 
    \begin{subfigure}[b]{0.49\textwidth}
        \centering
        \includegraphics[width=\linewidth]{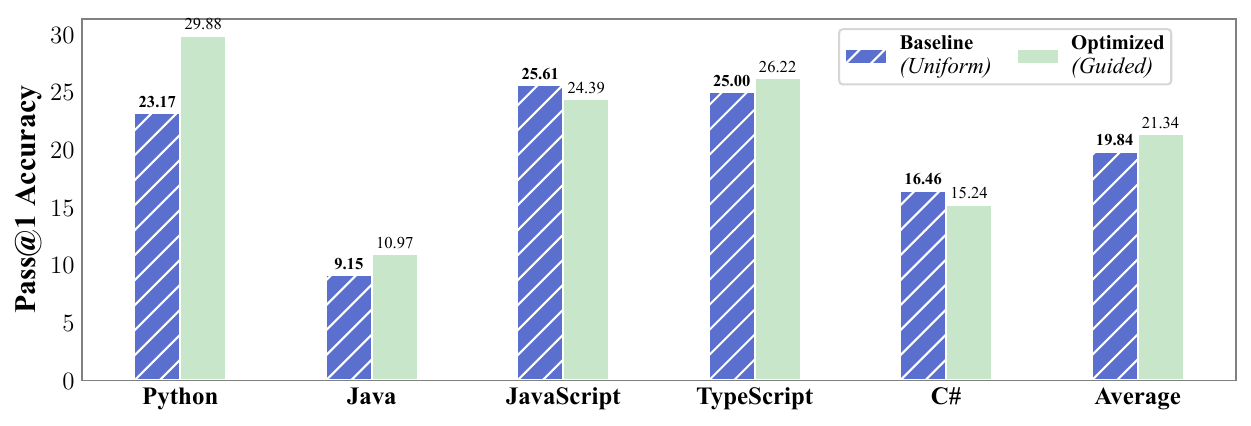} 
        \caption{Multilingual code generation performance}
        \label{fig:right_image}
    \end{subfigure}
    
    \vspace{-2mm}
    \caption{Pass@1 accuracy on the MultiPL-E benchmark and the BLEU scores of the code translation for baseline (uniform allocation) and optimized (guided allocation) strategies. Both LLMs are trained on 400B total tokens (350B code + 50B natural language text).}
    \label{tab:multiple_results}
\end{figure}

Both LLMs are evaluated on the multilingual code generation benchmark MultiPL-E across all $7$ PLs with the Pass@1 metric and our created code translation test set with the BLEU score. 
\autoref{tab:multiple_results} validates that the optimized allocation achieves higher average performance than uniform distribution under identical compute budgets. The results show that high-synergy pairs (e.g., JavaScript-TypeScript) benefit most from balanced allocation, Python improves with increased data due to high $\alpha_D$, while low-$\alpha_D$ languages (e.g., Rust) maintain strong performance despite reduced tokens. Importantly, no language suffers significant degradation, demonstrating that strategic reallocation finds a better equilibrium without creating imbalances.
Our results provide concrete evidence that multilingual scaling laws can directly inform data allocation strategies for each PL. 

\section{Related Work}
\paragraph{Scaling Laws} The systematic study of scaling laws in neural networks has evolved significantly since early observations~\cite{deep_learning_scaling_predictable,measure_effectes_data_parallellism} show power-law relationships between scale and performance. The foundational work~\cite{scaling_laws_lm} established that language model performance scales predictably with model size, dataset size, and compute, though this was significantly~\cite{chinchilla} in the Chinchilla paper, which demonstrated that models should be scaled equally in parameters and training tokens for compute-optimal training. This field has expanded to cover various domains, including transfer learning, data quality, and generative modeling beyond language, while the previous works ~\cite{emergent_ability_llm} documented emergent abilities that appear at certain scale thresholds, which is a phenomenon later challenged by potentially metric-dependent~\cite{are_emergent_abilities_llm}. The pioneering study about code scaling~\cite{code_scaling_law} establishes that code exhibits distinct and more data-hungry scaling laws than natural language, requiring a higher optimal data-to-parameter ratio but limited to the analysis of a single programming language.

\paragraph{Code Pre-training}
Recent advancements in code large language model (code LLM) pre-training have demonstrated remarkable progress in enabling models to understand, generate, and reason about programming languages~\cite{deepseek_coder,qwen25coder,starcoder,santacoder}. Early works, such as CodeBERT~\cite{code_bert} and GraphCodeBERT~\cite{graph_code_bert}, extend natural language pre-training techniques like masked language modeling and replaced token detection to source code, integrating both textual and structural representations through abstract syntax trees (ASTs). Subsequent efforts, including CodeT5~\cite{codet5} and CodeGen~\cite{codegen}, adopt encoder-decoder architectures for bidirectional and generative code tasks, leveraging large-scale datasets such as CodeSearchNet~\cite{codesearchnet}. More recent foundation models like Codex~\cite{codex}, StarCoder~\cite{starcoder}, Code Llama~\cite{codellama}, QwenCoder~\cite{qwen25coder}, and DeepSeek-Coder~\cite{deepseek_coder} have scaled pre-training to trillions of tokens across multiple programming languages, employing pre-training and post-training~\cite{coderl,deepseek_r1} to improve generalization, reasoning, and alignment with human intent.

\section{Conclusion}
In this work, we present the first systematic investigation of scaling laws for multilingual code pre-training, addressing critical gaps in our understanding of how programming language diversity affects model performance. Through over 1000+ experiments spanning multiple languages, model sizes, and dataset configurations, we establish several key findings: (1) different programming languages exhibit distinct scaling behaviors, with interpreted languages like Python showing larger scaling exponents than compiled languages like Rust, and intrinsic complexity ordering from C\# to Python reflected in their irreducible losses; (2) strategic language pairing during pre-training yields synergistic benefits, particularly for syntactically similar languages like Java-C\#, while most languages benefit from multilingual exposure compared to monolingual training; (3) parallel pairing strategies for organizing cross-lingual data significantly enhance translation capabilities with favorable scaling properties; and (4) our proportion-dependent multilingual scaling law enables optimal token allocation strategies that allocate more resources to high-utility languages and synergistic pairs while reducing allocation to fast-saturating languages, achieving superior average performance without degrading individual language capabilities. These findings provide actionable guidance for practitioners designing multilingual code models and establish a theoretical foundation for understanding cross-lingual transfer in code pre-training, ultimately enabling more efficient utilization of computational resources in training next-generation code LLMs.

\section*{Limitations}
While our work provides comprehensive insights into multilingual code scaling laws, several limitations warrant consideration. First, our experiments cover only seven programming languages, which, although representative of diverse paradigms, constitute a subset of production languages; extending findings to low-resource or domain-specific languages (e.g., SQL, assembly) remains unexplored. Second, our largest model reaches 14B parameters with 1T tokens, whereas state-of-the-art code LLMs exceed 100B parameters trained on multiple trillions of tokens (whether our scaling exponents hold at extreme scales requires validation). Third, our evaluation focuses on code translation and generation benchmarks, which may not fully capture performance on complex tasks like program repair or multi-file completion. Fourth, the synergy coefficients are fitted to our specific corpus; different data distributions may yield varying patterns requiring recalibration. Finally, our optimization assumes fixed token budgets and does not explore dynamic curriculum learning or adaptive sampling strategies that could further improve multilingual performance.

\clearpage

\bibliography{code_scalinglaw}

@misc{terminalbench,
      title={Terminal-Bench: A Benchmark for AI Agents in Terminal Environments}, 
      url={https://github.com/laude-institute/terminal-bench}, 
      author={The Terminal-Bench Team}, year={2025}, month={Apr}}

@Article{codellama,
  author       = {Roziere, Baptiste and Gehring, Jonas and Gloeckle, Fabian and Sootla, Sten and Gat, Itai and Tan, Xiaoqing Ellen and Adi, Yossi and Liu, Jingyu and Remez, Tal and Rapin, J{\'e}r{\'e}my and others},
  year         = {2023},
  journaltitle = {arXiv preprint arXiv:2308.12950},
  title        = {Code llama: Open foundation models for code},
}

@article{gpt4o,
  title={Gpt-4o system card},
  author={Hurst, Aaron and Lerer, Adam and Goucher, Adam P and Perelman, Adam and Ramesh, Aditya and Clark, Aidan and Ostrow, AJ and Welihinda, Akila and Hayes, Alan and Radford, Alec and others},
  journal={arXiv preprint arXiv:2410.21276},
  year={2024}
}

@article{qwen25coder,
  title={Qwen2. 5-Coder Technical Report},
  author={Hui, Binyuan and Yang, Jian and Cui, Zeyu and Yang, Jiaxi and Liu, Dayiheng and Zhang, Lei and Liu, Tianyu and Zhang, Jiajun and Yu, Bowen and Dang, Kai and others},
  journal={arXiv preprint arXiv:2409.12186},
  year={2024}
}

@inproceedings{code_bert,
  author       = {Zhangyin Feng and
                  Daya Guo and
                  Duyu Tang and
                  Nan Duan and
                  Xiaocheng Feng and
                  Ming Gong and
                  Linjun Shou and
                  Bing Qin and
                  Ting Liu and
                  Daxin Jiang and
                  Ming Zhou},
  title        = {CodeBERT: {A} Pre-Trained Model for Programming and Natural Languages},
  booktitle    = {Findings of the Association for Computational Linguistics: {EMNLP}
                  2020, Online Event, 16-20 November 2020},
  series       = {Findings of {ACL}},
  volume       = {{EMNLP} 2020},
  pages        = {1536--1547},
  publisher    = {Association for Computational Linguistics},
  year         = {2020},
  url          = {https://doi.org/10.18653/v1/2020.findings-emnlp.139},
  doi          = {10.18653/V1/2020.FINDINGS-EMNLP.139},
}

@article{deepseek_coder,
  title={DeepSeek-Coder: When the Large Language Model Meets Programming--The Rise of Code Intelligence},
  author={Guo, Daya and Zhu, Qihao and Yang, Dejian and Xie, Zhenda and Dong, Kai and Zhang, Wentao and Chen, Guanting and Bi, Xiao and Wu, Y and Li, YK and others},
  journal={arXiv preprint arXiv:2401.14196},
  year={2024},
  url={https://arxiv.org/abs/2401.14196},
}

@article{gpt3,
  title={Language models are few-shot learners},
  author={Brown, Tom and Mann, Benjamin and Ryder, Nick and Subbiah, Melanie and Kaplan, Jared D and Dhariwal, Prafulla and Neelakantan, Arvind and Shyam, Pranav and Sastry, Girish and Askell, Amanda and others},
  journal={Advances in neural information processing systems},
  volume={33},
  pages={1877--1901},
  year={2020}
}

@article{starcoder,
  author       = {Raymond Li and
                  Loubna Ben Allal and
                  Yangtian Zi and
                  Niklas Muennighoff and
                  Denis Kocetkov and
                  Chenghao Mou and
                  Marc Marone and
                  Christopher Akiki and
                  Jia Li and
                  Jenny Chim and
                  Qian Liu and
                  Evgenii Zheltonozhskii and
                  Terry Yue Zhuo and
                  Thomas Wang and
                  Olivier Dehaene and
                  Mishig Davaadorj and
                  Joel Lamy{-}Poirier and
                  Jo{\~{a}}o Monteiro and
                  Oleh Shliazhko and
                  Nicolas Gontier and
                  Nicholas Meade and
                  Armel Zebaze and
                  Ming{-}Ho Yee and
                  Logesh Kumar Umapathi and
                  Jian Zhu and
                  Benjamin Lipkin and
                  Muhtasham Oblokulov and
                  Zhiruo Wang and
                  Rudra Murthy V and
                  Jason Stillerman and
                  Siva Sankalp Patel and
                  Dmitry Abulkhanov and
                  Marco Zocca and
                  Manan Dey and
                  Zhihan Zhang and
                  Nour Moustafa{-}Fahmy and
                  Urvashi Bhattacharyya and
                  Wenhao Yu and
                  Swayam Singh and
                  Sasha Luccioni and
                  Paulo Villegas and
                  Maxim Kunakov and
                  Fedor Zhdanov and
                  Manuel Romero and
                  Tony Lee and
                  Nadav Timor and
                  Jennifer Ding and
                  Claire Schlesinger and
                  Hailey Schoelkopf and
                  Jan Ebert and
                  Tri Dao and
                  Mayank Mishra and
                  Alex Gu and
                  Jennifer Robinson and
                  Carolyn Jane Anderson and
                  Brendan Dolan{-}Gavitt and
                  Danish Contractor and
                  Siva Reddy and
                  Daniel Fried and
                  Dzmitry Bahdanau and
                  Yacine Jernite and
                  Carlos Mu{\~{n}}oz Ferrandis and
                  Sean Hughes and
                  Thomas Wolf and
                  Arjun Guha and
                  Leandro von Werra and
                  Harm de Vries},
  title        = {{StarCoder}: May the source be with you!},
  journal      = {arXiv preprint arXiv:2305.06161},
  volume       = {abs/2305.06161},
  year         = {2023},
  url          = {https://doi.org/10.48550/arXiv.2305.06161},
  doi          = {10.48550/arXiv.2305.06161},
  eprinttype    = {arXiv},
  eprint       = {2305.06161},
}

@article{chinchilla,
  title={Training compute-optimal large language models},
  author={Hoffmann, Jordan and Borgeaud, Sebastian and Mensch, Arthur and Buchatskaya, Elena and Cai, Trevor and Rutherford, Eliza and Casas, Diego de Las and Hendricks, Lisa Anne and Welbl, Johannes and Clark, Aidan and others},
  journal={arXiv preprint arXiv:2203.15556},
  year={2022}
}

@article{codex,
  author       = {Mark Chen and
                  Jerry Tworek and
                  Heewoo Jun and
                  Qiming Yuan and
                  Henrique Pond{\'{e}} de Oliveira Pinto and
                  Jared Kaplan and
                  Harrison Edwards and
                  Yuri Burda and
                  Nicholas Joseph and
                  Greg Brockman and
                  Alex Ray and
                  Raul Puri and
                  Gretchen Krueger and
                  Michael Petrov and
                  Heidy Khlaaf and
                  Girish Sastry and
                  Pamela Mishkin and
                  Brooke Chan and
                  Scott Gray and
                  Nick Ryder and
                  Mikhail Pavlov and
                  Alethea Power and
                  Lukasz Kaiser and
                  Mohammad Bavarian and
                  Clemens Winter and
                  Philippe Tillet and
                  Felipe Petroski Such and
                  Dave Cummings and
                  Matthias Plappert and
                  Fotios Chantzis and
                  Elizabeth Barnes and
                  Ariel Herbert{-}Voss and
                  William Hebgen Guss and
                  Alex Nichol and
                  Alex Paino and
                  Nikolas Tezak and
                  Jie Tang and
                  Igor Babuschkin and
                  Suchir Balaji and
                  Shantanu Jain and
                  William Saunders and
                  Christopher Hesse and
                  Andrew N. Carr and
                  Jan Leike and
                  Joshua Achiam and
                  Vedant Misra and
                  Evan Morikawa and
                  Alec Radford and
                  Matthew Knight and
                  Miles Brundage and
                  Mira Murati and
                  Katie Mayer and
                  Peter Welinder and
                  Bob McGrew and
                  Dario Amodei and
                  Sam McCandlish and
                  Ilya Sutskever and
                  Wojciech Zaremba},
  title        = {Evaluating Large Language Models Trained on Code},
  journal      = {arXiv preprint arXiv:2107.03374},
  volume       = {abs/2107.03374},
  year         = {2021},
  url          = {https://arxiv.org/abs/2107.03374},
  eprinttype    = {arXiv},
  eprint       = {2107.03374},
}

@article{santacoder,
  title={{SantaCoder}: Don't reach for the stars!},
  author={Allal, Loubna Ben and Li, Raymond and Kocetkov, Denis and Mou, Chenghao and Akiki, Christopher and Ferrandis, Carlos Munoz and Muennighoff, Niklas and Mishra, Mayank and Gu, Alex and Dey, Manan and others},
  journal={arXiv preprint arXiv:2301.03988},
  year={2023},
  url={https://arxiv.org/abs/2301.03988}
}

@article{codet5,
  title={{CodeT5}: Identifier-aware unified pre-trained encoder-decoder models for code understanding and generation},
  author={Wang, Yue and Wang, Weishi and Joty, Shafiq and Hoi, Steven CH},
  journal={arXiv preprint arXiv:2109.00859},
  year={2021},
  url={https://arxiv.org/abs/2109.00859},
}

@inproceedings{codegen,
  author       = {Erik Nijkamp and
                  Bo Pang and
                  Hiroaki Hayashi and
                  Lifu Tu and
                  Huan Wang and
                  Yingbo Zhou and
                  Silvio Savarese and
                  Caiming Xiong},
  title        = {{CodeGen}: An Open Large Language Model for Code with Multi-Turn Program
                  Synthesis},
  booktitle    = {The Eleventh International Conference on Learning Representations,
                  {ICLR} 2023, Kigali, Rwanda, May 1-5, 2023},
  publisher    = {OpenReview.net},
  year         = {2023},
}

@article{codesearchnet,
  author       = {Hamel Husain and
                  Ho{-}Hsiang Wu and
                  Tiferet Gazit and
                  Miltiadis Allamanis and
                  Marc Brockschmidt},
  title        = {CodeSearchNet Challenge: Evaluating the State of Semantic Code Search},
  journal      = {arXiv preprint arXiv:1909.09436},
  volume       = {abs/1909.09436},
  year         = {2019},
  url          = {http://arxiv.org/abs/1909.09436},
  eprinttype    = {arXiv},
  eprint       = {1909.09436},
}

@article{deepseek_r1,
  title={Deepseek-r1: Incentivizing reasoning capability in llms via reinforcement learning},
  author={Guo, Daya and Yang, Dejian and Zhang, Haowei and Song, Junxiao and Zhang, Ruoyu and Xu, Runxin and Zhu, Qihao and Ma, Shirong and Wang, Peiyi and Bi, Xiao and others},
  journal={arXiv preprint arXiv:2501.12948},
  year={2025}
}

@misc{cursor2025features,
  title={Cursor Features},
  author={{Anysphere Inc.}},
  howpublished={\url{https://cursor.com/features}},
  year={2025}
}

@article{scaling_laws_lm,
  title={Scaling laws for neural language models},
  author={Kaplan, Jared and McCandlish, Sam and Henighan, Tom and Brown, Tom B and Chess, Benjamin and Child, Rewon and Gray, Scott and Radford, Alec and Wu, Jeffrey and Amodei, Dario},
  journal={arXiv preprint arXiv:2001.08361},
  year={2020}
}

@article{deepseekv1,
  title={Deepseek llm: Scaling open-source language models with longtermism},
  author={Bi, Xiao and Chen, Deli and Chen, Guanting and Chen, Shanhuang and Dai, Damai and Deng, Chengqi and Ding, Honghui and Dong, Kai and Du, Qiushi and Fu, Zhe and others},
  journal={arXiv preprint arXiv:2401.02954},
  year={2024}
}

@article{code_scaling_law,
  title={Scaling Laws for Code: A More Data-Hungry Regime},
  author={Luo, Xianzhen and Zheng, Wenzhen and Zhu, Qingfu and Zhang, Rongyi and Li, Houyi and Huang, Siming and Fan, YuanTao and Che, Wanxiang},
  journal={arXiv preprint arXiv:2510.08702},
  year={2025}
}

@article{deep_learning_scaling_predictable,
  title={Deep learning scaling is predictable, empirically},
  author={Hestness, Joel and Narang, Sharan and Ardalani, Newsha and Diamos, Gregory and Jun, Heewoo and Kianinejad, Hassan and Patwary, Md Mostofa Ali and Yang, Yang and Zhou, Yanqi},
  journal={arXiv preprint arXiv:1712.00409},
  year={2019}
}

@article{measure_effectes_data_parallellism,
  title={Measuring the effects of data parallelism on neural network training},
  author={Shallue, Christopher J and Lee, Jaehoon and Antognini, Joseph and Sohl-Dickstein, Jascha and Frostig, Roy and Dahl, George E},
  journal={Journal of Machine Learning Research},
  volume={20},
  number={112},
  pages={1--49},
  year={2019}
}

@article{emergent_ability_llm,
  title={Emergent abilities of large language models},
  author={Wei, Jason and Tay, Yi and Bommasani, Rishi and Raffel, Colin and Zoph, Barret and Borgeaud, Sebastian and Yogatama, Dani and Bosma, Maarten and Zhou, Denny and Metzler, Donald and others},
  journal={arXiv preprint arXiv:2206.07682},
  year={2022}
}

@article{are_emergent_abilities_llm,
  title={Are emergent abilities of large language models a mirage?},
  author={Schaeffer, Rylan and Miranda, Brando and Koyejo, Sanmi},
  journal={Advances in neural information processing systems},
  volume={36},
  pages={55565--55581},
  year={2023}
}

@inproceedings{graph_code_bert,
  author       = {Daya Guo and
                  Shuo Ren and
                  Shuai Lu and
                  Zhangyin Feng and
                  Duyu Tang and
                  Shujie Liu and
                  Long Zhou and
                  Nan Duan and
                  Alexey Svyatkovskiy and
                  Shengyu Fu and
                  Michele Tufano and
                  Shao Kun Deng and
                  Colin B. Clement and
                  Dawn Drain and
                  Neel Sundaresan and
                  Jian Yin and
                  Daxin Jiang and
                  Ming Zhou},
  title        = {GraphCodeBERT: Pre-training Code Representations with Data Flow},
  booktitle    = {9th International Conference on Learning Representations, {ICLR} 2021,
                  Virtual Event, Austria, May 3-7, 2021},
  publisher    = {OpenReview.net},
  year         = {2021},
  url          = {https://openreview.net/forum?id=jLoC4ez43PZ},
  bibsource    = {dblp computer science bibliography, https://dblp.org}
}

@inproceedings{coderl,
  author       = {Hung Le and
                  Yue Wang and
                  Akhilesh Deepak Gotmare and
                  Silvio Savarese and
                  Steven Chu{-}Hong Hoi},
  editor       = {Sanmi Koyejo and
                  S. Mohamed and
                  A. Agarwal and
                  Danielle Belgrave and
                  K. Cho and
                  A. Oh},
  title        = {CodeRL: Mastering Code Generation through Pretrained Models and Deep
                  Reinforcement Learning},
  booktitle    = {Advances in Neural Information Processing Systems 35: Annual Conference
                  on Neural Information Processing Systems 2022, NeurIPS 2022, New Orleans,
                  LA, USA, November 28 - December 9, 2022},
  year         = {2022},
  url          = {http://papers.nips.cc/paper\_files/paper/2022/hash/8636419dea1aa9fbd25fc4248e702da4-Abstract-Conference.html},
  bibsource    = {dblp computer science bibliography, https://dblp.org}
}

@article{osagents,
  title={Os agents: A survey on mllm-based agents for general computing devices use},
  author={Hu, Xueyu and Xiong, Tao and Yi, Biao and Wei, Zishu and Xiao, Ruixuan and Chen, Yurun and Ye, Jiasheng and Tao, Meiling and Zhou, Xiangxin and Zhao, Ziyu and others},
  journal={arXiv preprint arXiv:2508.04482},
  year={2025}
}

@article{guiagents_foundation_models,
  title={Gui agents with foundation models: A comprehensive survey},
  author={Wang, Shuai and Liu, Weiwen and Chen, Jingxuan and Zhou, Yuqi and Gan, Weinan and Zeng, Xingshan and Che, Yuhan and Yu, Shuai and Hao, Xinlong and Shao, Kun and others},
  journal={arXiv preprint arXiv:2411.04890},
  year={2024}
}

\end{document}